\documentclass[letterpaper, 10 pt, conference]{ieeeconf}  % Comment this line out
\IEEEoverridecommandlockouts                              % This command is only
\usepackage{amssymb}
\usepackage{graphicx,amsmath}
\usepackage{lineno}
\usepackage{caption,url}
\usepackage{float}

\usepackage{subcaption}
\usepackage{gensymb}
\usepackage[colorlinks]{hyperref}
\hypersetup{
    colorlinks,
    linkcolor=black,
    citecolor=black,
    filecolor=black,
    urlcolor=black,
}
\title{\LARGE \bf
Gait Library Synthesis for Quadruped Robots via \\
Augmented Random Search
}

\author{Sashank Tirumala, Aditya Sagi, Kartik Paigwar, Ashish Joglekar, \\ Shalabh Bhatnagar, Ashitava Ghosal, Bharadwaj Amrutur, Shishir Kolathaya % <-this % stops a space
\thanks{*This work is supported by DST INSPIRE Fellowship IFA17-ENG212, and Robert Bosch Center for Cyber-Physical Systems, IISc, Bengaluru.}% <-this % stops a space
\thanks{ S. Tirumala is with the Department of Engineering Design at the Indian Institute of Technology-Madras, Chennai, A. Sagi, K. Paigwar and A. Joglekar are with the Robert Bosch Center for Cyber-Physical Systems at the Indian Institute of Science, Bangalore}
\thanks{S. Kolathaya is an INSPIRE Faculty Fellow and S. Bhatnagar, A. Ghosal, B. Amrutur are with the faculty of the Robert Bosch Center for Cyber-Physical Systems, Indian Institute of Science, Bengaluru, India. {\tt\small email: \{shishirk,shalabh,asitava,amrutur\} at iisc.ac.in}}%
}%

\begin{document}

\maketitle
\thispagestyle{empty}
\pagestyle{empty}

%%%%%%%%%%%%%%%%%%%%%%%%%%%%%%%%%%%%%%%%%%%%%%%%%%%%%%%%%%%%%%%%%%%%%%%%%%%%%%%%
\begin{abstract}

In this paper, with a view toward fast deployment of learned locomotion gaits in low-cost hardware, we generate a library of walking trajectories, namely, forward trot, backward trot, side-step, and turn in our custom built quadruped robot, Stoch 2, using reinforcement learning.
There are existing approaches that determine optimal policies for each time step, whereas we determine an optimal policy, in the form of end-foot trajectories, for each half walking step i.e., swing phase and stance phase. 
 The way-points for the foot trajectories are obtained from a linear policy, i.e., a linear function of the states of the robot, and cubic splines are used to interpolate between these points.
Augmented Random Search, a model-free and gradient-free learning algorithm, is used to learn the policy in simulation. This learned policy is then deployed on hardware, yielding a trajectory in every half walking step. 
Different locomotion patterns are learned in simulation by enforcing a preconfigured phase shift between the trajectories of different legs.
Transition from one gait to another is achieved by using a low-pass filter for the phase, and the sim-to-real transfer is improved by a linear transformation of the states obtained through regression.

% This is due to the fact that the basic configuration patterns used for locomotion are periodic and low dimensional. This is similar to the 

% There are existing results on a trotting behaviors via the deployment of deep neural networks (DNN) based reinforcement learning algorithms, while our approach uses a linear policy, and this policy is learned via a model-free, gradient-free learning algorithm, augmented random search (ARS).

%  In addition, to address the sim-to-real transfer problem, by obtaining a linear transformation of the learned policy via regression.

% This issue is known in the literature as the Sim-to-Real gap.

\end{abstract}


%%%%%%%%%%%%%%%%%%%%%%%%%%%%%%%%%%%%%%%%%%%%%%%%%%%%%%%%%%%%%%%%%%%%%%%%%%%%%%%%
\section{Introduction}

Quadrupedal locomotion with multiple types of gaits has been successfully demonstrated via numerous techniques---inverted pendulum models \cite{raibert1986legged}, zero-moment point \cite{vukobratovic2004zero}, hybrid zero dynamics \cite{hzd_grizzle}, and central pattern generators (CPG) \cite{ijspeert2008central}---to name a few. 
However, due to the increase in computational resources, data driven approaches like reinforcement learning (RL) \cite{kober2013reinforcement} are gaining popularity today. From a user's point of view, RL only requires the determination of a scalar reward, like distance travelled, energy consumed etc, and then the algorithm identifies the best controller for walking by itself.
The success of this methodology was enabled by the use of deep neural networks in RL, popularly known as deep reinforcement learning (D-RL) \cite{google_paper,Hwangboeaau5872}.

The considerable success shown by D-RL are mostly for simulated robotic tasks \cite{kober2013reinforcement} even today. Translating these results in real hardware are faced with numerous challenges. 
The use of deep neural networks (DNN) entails a large computational overhead to obtain an inference. This not only requires expensive hardware, but also results in higher power consumption, which may not be economical for a large commercial deployment of robots.
In addition, D-RL based techniques require a significant number of training iterations (to the order of millions), which are detrimental to hardware. 
A possible alternative would be to train in a simulated environment, and then deploy on hardware. This was demonstrated by \cite{google_paper}, \cite{Hwangboeaau5872}, \cite{xie2019iterative} for quadrupeds and bipeds, where the neural networks used sensory feedback to control the motor joints in real-time. However, this type of sim-to-real transfer presents with itself several challenges. Some of these challenges are mainly attributed to the modeling uncertainty, sensory noise, control saturations etc.

\begin{figure}[t!]
\centering
\vspace{2mm}
\includegraphics[width = \linewidth] {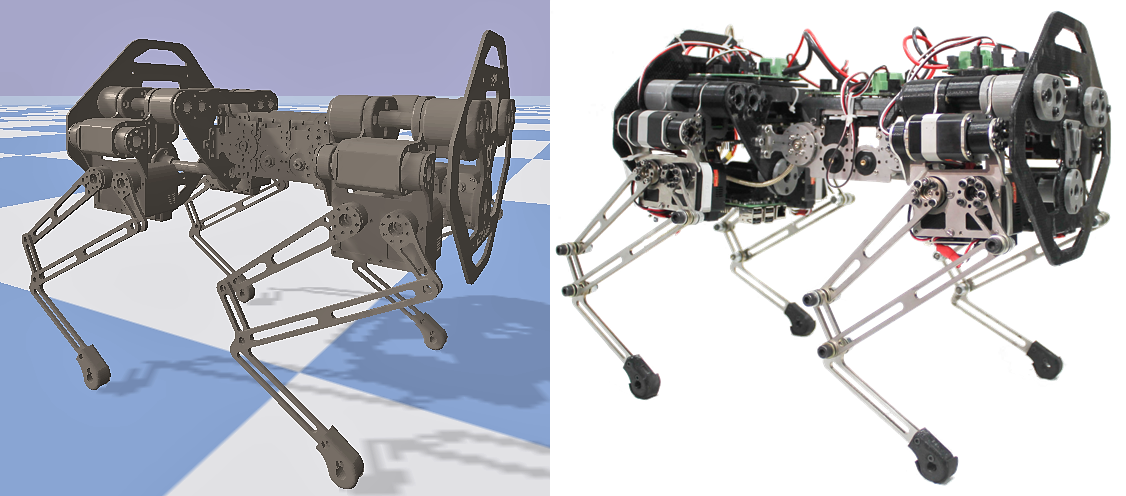}
\caption{Figure showing the custom built quadruped robot, Stoch 2. Simulated version is shown in the left, and the actual hardware is shown in the right.}
\label{fig:PyBullet}
\end{figure}

%Researchers have tried to eliminate the extensive manual tuning and expertise required by using evolutionary strategies and reinforcement learning  \cite{singla2018realizing}.  
It is worth noting that some of the solutions to the above presented challenges were addressed in part via the traditional RL techniques \cite{Hornby1999Autonomous,quinlan2003techniques,kohl2004policy,tedrake2004stochastic}. In fact, RL was used as an optimization tool (gait generating tool) by most of the teams participating in the quadruped RoboCup soccer league in 2004 as stated by \cite{Chalup2007Machine}. These techniques have been used to obtain the parameterized  foot trajectories that can be deployed on the robot. This can be implemented in low cost hardware and requires no extra computing power. However, these works implemented open-loop controllers to playback the trajectories on the robots, thereby, limiting their capability to handle external disturbances due to the lack of feedback in the system.

Similar to the methodologies followed in the past in RL, we would like to realize locomotion on a low-cost embedded platform and, at the same time, incorporate some form of feedback that will help improve robustness in real-time. By limiting our controller to a linear policy, i.e., a single matrix with no activation functions, our goal is to learn this policy with the modern tools available for learning. We chose to use a model-free and gradient-free algorithm, Augmented Random Search (ARS), which was shown to be at least as sample efficient as DNN based policy gradient algorithms \cite{mania2018simple} existing in literature. 
We present a control architecture that combines a trajectory-based framework with RL to obtain a controller that is capable of handling external disturbances and exhibiting multiple complex quadruped locomotion behaviours while requiring little manual tuning. 

\subsection{Related work on ARS}
ARS was successfully demonstrated on a variety of robotic control tasks (including quadrupedal locomotion) with a linear policy \cite{mania2018simple}. These types of policies have also been deployed successfully on the quadruped robot Minitaur \cite{PMTG}. In particular, \cite{PMTG} used a policy modulating trajectory generator (PMTG) that modulated a trajectory and simultaneously modified the outputs to obtain the final commands for the motors. This was successfully used with linear policies to achieve walking and bounding gaits. \cite{jain2019hierarchical} extended the PMTG framework to include turning gaits in Minitaur. However, the turning demonstrated by Minitaur was limited due to the absence of abduction motors. In addition, the PMTG framework updates the angle commands after every time step. Therefore, with a view toward lowering the computational overhead in hardware, and including abduction based gaits like side-stepping, in-place turning, we make a slight deviation from this approach, by directly obtaining the end-foot trajectories from a trained policy. The policy is updated every half-step, and the motors are commanded to track the resulting trajectories. We demonstrate a library of gaits forward trot, backward trot, side step and turn, in the custom built quadruped robot Stoch $2$. Towards the end we demonstrate robustness by showing that the linear policy is able to reject disturbances such as pushing and pressing.

\subsection{Organization}

The paper is structured as follows: Section \ref{sec:background} will provide a brief background on reinforcement learning (RL) along with a description of the robot hardware and the associated control framework, Section \ref{sec:learningalgorithm} will describe the methodology used for training followed by the experimental results in Section \ref{sec:experiment}.

%\textbf{fix this, also refer to the sections by the references}This paper is divided as follows Section 3 provides the needed background in Reinforcement Learning, Central Pattern Generators, Active splines and the robot hardware. Section 4 describes our controller architecture and methodology, Section 5,6,7 describe our results experiments and conclusions.

\section{Robot model and control}
\label{sec:background}

In this section, we will discuss the reinforcement learning framework used for our custom built quadruped robot Stoch 2. Specifically, we will provide details about the hardware, the associated model, and the trajectory tracking control framework used for the various gaits of the robot.

\subsection{Hardware description of Stoch $2$}
\textit{Stoch $2$}  is  a  quadruped  robot designed  and developed in-house at the Indian Institute of Science (IISc), Bengaluru, India. 
It is a second generation robot in the \textit{Stoch} series  \cite{singla2018realizing, dhaivatdesigndevelopment}. It is similar to \textit{Stoch} \cite{dhaivatdesigndevelopment} in form factor, and weighs approximately $4$kg. Each leg contains three joints---hip flexion/extension, hip abduction and knee flexion/extension. Servo motors from Kondo Kogaku (model: KRS6003) are used to actuate these joints. The main controller on the robot is an ARM processor (the Tiva TM4C123GH6PM) running at $80$~MHz.
This controller is placed on the central module that connects the front and back parts of Stoch $2$. The URDF model used in the simulator was created directly from the SolidWorks assembly (see Fig. \ref{fig:PyBullet}). Overall, the robot simulation model consists of $6$ base degrees-of-freedom and $12$ actuated degrees-of-freedom.

\subsection{Kinematic description}\label{subsec:kinematic}

Each leg comprises of a parallel five-bar linkage mechanism where two of the links are actuated as shown in Fig. \ref{fig:IK_S2}. This enables the end foot point to follow the given trajectory in a plane. The two actuators which control the motion of upper hip and knee linkages are mounted on a fixed link. These actuated linkages in turn connect to the lower linkages via revolute joints.

In this paper, we focus on realizing trajectories of the feet in polar coordinates (see Section \ref{subsec:actionspace} ahead). 
As seen from Fig \ref{fig:IK_S2}, the five-bar linkage is divided into two serial 2-R linkages and solved for each branch. The details of the equations for a serial 2-R linkage can be found in \cite{dhaivatdesigndevelopment}.

\subsection{Reinforcement learning for walking}
We formulate the problem of locomotion as a Markov Decision Process (MDP). Here an MDP is a 5-tuple $\{S, A, r, P, \gamma \}$ where $S \subset \mathbb{R}^n$ refers to set of states of the robot, $A \subset \mathbb{R}^m$ refers to the set of actions, $r : S \times A \rightarrow \mathbb{R}$ refers to the reward received for every $(S,A)$  pair, $P : S\times A \times S \rightarrow [0,1]$ refers to the transition probabilities between two states for a given action, and $\gamma \in (0,1)$ is the discount factor of the MDP. Description of the states, actions for Stoch 2 are explained later on in this section. Rewards are chosen depending upon the gait, which are explained in Section \ref{subsec:training}. Given a policy $\pi : S \to A$, we evaluate it for an episode. In this formulation, the optimal policy is the policy that maximizes the return ($R$):
\begin{align}
R = \mathbb{E}[r_t+ \gamma r_{t+1} + \gamma^2 r_{t+2} + \dots],
\end{align}
where the subscript for $r_t$ denotes the step index. Since walking is periodic in nature, each step here corresponds to one cycle/loop of the foot trajectory. Therefore, our policy is evaluated two times for every gait cycle/loop of the robot. Henceforth, we will refer to this half-cycle by ``gait step". We will describe the trajectories that yield these gait steps next.

\subsection{Cubic splines}
A cubic Hermite spline is a piece-wise third degree polynomial, which not only interpolates the values between two points, but also their derivatives.
The equation of a Hermite spline specified by its endpoints $\{w_0,w_1$\} and tangents $\{w'_0, w'_1\}$ is
\begin{align}
w(t) = 
  \left[ {\begin{array}{c}
   2t^3 - 3t^2 +1 \\
   -2t^3 +3t^2 \\
   t^3-2t^2 +t \\
   t^3 -t^2
  \end{array} } \right]^T \times
  \left[ {\begin{array}{c}
   w_0 \\
   w_1 \\
   w'_0 \\
   w'_1
  \end{array} } \right],
\end{align}
where $t$ is the phasing variable. Therefore, given $n$ points $(w_0,t_0), (w_1, t_1), \dots, (w_{n-1}, t_{n-1})$, we can connect the adjacent pair of points resulting in a loop. Since the tangents are user definable, we choose them to be:
\begin{align}
    w'_i = \frac{w_{i+1} - w_{i-1}}{t_{i+1} - t_{i-1}},
\end{align}
and if $i=0$, we choose $w_{-1} = w_{n-1}$.
% Directly using the above formulation may cause the learnt policy to generate gaits that have loops, cusps and high accelerations that damage the motor hardware. To prevent such gaits, we use a modification of the Hermite spline called the centripetal Catmull-Rom Spline \textbf{cite a paper} that provably avoid loops or cusps. In addition, these splines provide smooth motions in the motors by limiting the accelerations (second derivatives). Given four points in a plane $(w_0, t_0), (w_1, t_1), (w_2, w_2), (w_3, w_3)$, the derivatives (denoted by $w_i'$) at the end points are given by:
% %In this formulation the tangents at the end points $\{x_1,x_2$\} given points : $\{x_0,x_1,x_2,x_3$\} are given by [Let $y = p(x)$] :
% $$w'_0 = (t_1 -t_0) *\left (\frac{w_1 -w_0}{t_1 -t_0} - \frac{w_2 -w_0}{t_2 - t_0} + \frac{w_2 -w_1}{t_2 - t_1} \right) $$
% $$w'_1 = (t_1 -t_0) * \left (\frac{w_2 -w_1}{t_2 -t_1} - \frac{w_3 -w_1}{t_3 - t_1} + \frac{w_3 -w_2}{t_3 - t_2} \right). $$
% We obtain a polynomial that connect these four points. In a similar fashion, we obtain the next polynomial for the next quartet of points: $(w_1, t_1), (w_2, t_2), (w_3, t_3), (w_4, t_4)$. This is repeated until the loop is closed i.e., by enforcing the last point $(w_n,w_n) = (t_1,t_1)$. 
It is worth noting that depending upon the gait, the plane containing these points is oriented appropriately w.r.t. the foot, and the corresponding joints angle trajectories are obtained via inverse kinematics. %For example, the $n$ points are placed on the $x-y$ plane for trotting, and $y-z$ plane for side-stepping. 
See Fig. \ref{fig:IKTrajspline} and 
Section \ref{subsec:training} for more details.

\begin{figure}
    \centering
    \includegraphics[width = \linewidth]{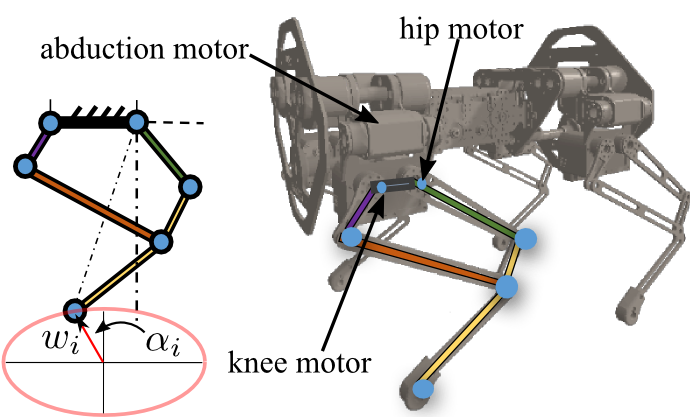}
    \caption{A five-bar mechanism is used as the legs of the quadruped robot. This mechanism is actuated by the motors located at the main torso of the robot.}
    \label{fig:IK_S2}
\end{figure}

\subsection{State space}

The state is represented by motor angles of the robot.
Hence our state space is a $12$ dimensional vector space: $S \subset \mathbb{R}^{12}$. Note that other sensory values such as angular velocities, accelerations and currents were ignored due to the fact that the policy was updated after every gait step.

\subsection{Action space}
\label{subsec:actionspace}
\begin{figure}
    \centering
    \includegraphics[width = \linewidth]{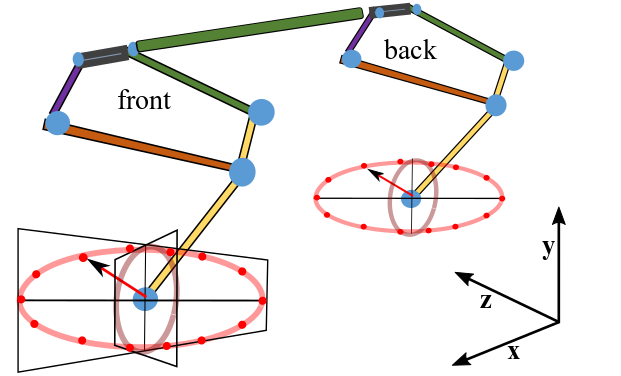}
    \caption{Figure showing the construction of the splines for the front and back legs of the robot. 
    Two types of ellipses are shown each on the $x-y$ and $y-z$ plane respectively. These ellipses are drawn to indicate the plane on which the splines are constructed.
    For illustration, the control points are shown in the form of red dots on one of the ellipses. For trotting, the $x-y$ plane is used, and for side-stepping, the $y-z$ plane is used. For turning, the planes are oriented at $+/- 45^\circ$ yaw w.r.t. the $x$ axis were used.}
    \label{fig:IKTrajspline}
\end{figure}

In reference \cite{action_space}, the actuator joint angles were selected for the action space. In reference \cite{singla2018realizing}, the legs' end-point positions in polar coordinates were selected for the action space. The polar coordinates for the four legs collectively provided an $8$-dimensional action space. In this paper, since we are using splines to determine the end-foot trajectories, we require the actions to yield the control points for these splines (see Fig. \ref{fig:IKTrajspline}). We express the splines in polar coordinates with the center located directly under the hip for each leg. Let $w_i$ denote the radius, and $\alpha_i$ (used in place of $t$) denote the phase angle (see Figs. \ref{fig:IK_S2}, \ref{fig:IKTrajspline}). Accordingly, we have 
$(w_0,\alpha_0), \dots, (w_{n-1},\alpha_{n-1})$, where the $\alpha_i$'s are obtained by evenly spacing from $0$ to $2\pi$~rad, and $w_i$'s are obtained from a policy. Note that $w_i$'s are updated as a function of $s$ after every gait step i.e., whenever the phase crosses $0$, $\pi$.

Having obtained the end-foot trajectory from the control points, the joint angles are obtained via an inverse kinematics solver. More details about the inverse kinematics for parallel $5$-bar linkages are provided in \cite{dhaivatdesigndevelopment}. Inclusion of more points allow the spline to take more complex shapes. It was empirically observed that $6$ control points were incapable of generating stable gaits, and after $18$ points, there was not much improvement in the gaits obtained. 
It is worth noting that the plane on which the control points are placed is predetermined depending upon the gait (more details are given in Section \ref{subsec:training}). In addition, with different phase shifts and signs, we run the same trajectory on all the four legs.
Therefore, we choose the action space to be an $18$ dimensional vector that is constrained to remain within the leg's work-space using a bounding box.

\section{Training and Simulation}
\label{sec:learningalgorithm}
Having defined the model and the control methodology, we are now ready to discuss the policy and training algorithm used for Stoch 2.

\subsection{Algorithm}
Since the goal is to realize a library of gaits in low-cost hardware, we need a simple representation of the policy, one that is capable of running on a single embedded Tiva-M Series Microcontroller that contains only one floating point unit. We chose the simplest representation, a linear policy, thus our policy is a matrix that multiplies with the state at every step to output the action. Augmented Random Search (ARS) \cite{mania2018simple}, is a learning algorithm which is designed for finding linear deterministic policies, and is known to be on par with other model-free Reinforcement Learning algorithms for robot control tasks. In literature, this algorithm was successfully demonstrated on the simulated MuJoCo robot control tasks to obtain rewards that were on par with the rewards obtained by deep nets trained using traditional Reinforcement Learning Algorithms like TRPO and PPO \cite{mania2018simple}. Our experiments on the Stoch2 environment also demonstrate the same %[TODO : ADD GRAPH OR TABLE]
% Policy Gradient Algorithms rely upon optimizing parameterized policies with respect to the expected return by gradient descent. 
Therefore we choose the policy to be $\pi(s):=M s$, where $M\in \mathbb{R}^{18\times 12}$ is the matrix that maps the $12$ motor angles to the $18$ control points. Let the parameters of $M$ be denoted by $\theta$. In this case as we are not adding any constraints to the matrix $M$, the parameters $\theta$ of $M$ are simply each element of $M$. Then the goal of ARS is to determine the $\theta$, of the matrix $M$, that yields the best rewards which in turn leads to the best locomotion gaits for Stoch 2.

ARS is a policy gradient algorithm, i.e, it optimizes a parameterized policy by moving along the gradient of the function that maps the parameters of the policy to the expected reward. Since the function is stochastic in nature, different algorithms use different estimates of the gradient. ARS estimates the gradient through a finite difference method as opposed to likelihood ratio methods used in other algorithms like PPO \cite{PPO} or TRPO \cite{TRPO}. We use Version $\textbf{V-1t}$ of ARS from \cite{mania2018simple}. $\textbf{V-1t}$ performs the gradient descent step without normalization of state or action space as in $\textbf{V-2}$, and averages a subset of the top performing directions to determine the final direction of the gradient along which the policy should move. We do not require normalization of state space since all the motor angles vary between the same limits. In addition, we move along the average of a subset of top performing directions $\textbf{V-1t}$, unlike in $\textbf{V-1}$, which uses the average of all directions. This speeds up the training process by ignoring poor directions. Since a properly tuned $\textbf{V-1t}$ includes the possibility of choosing the set of all directions, $\textbf{V-1t}$ can not perform worse than $\textbf{V-1}$. In particular, we pick $N$ i.i.d. directions $\{\delta_i\}_{\{i = 1,\dots,N\}}$ from a normal distribution, where $N$ is the dimension of the policy parameters, i.e., $\theta \in \mathbb{R}^{N}$. In our case, $N=18\times 12$, since we have $18$ actions and $12$ states. With a scaling factor of $\nu>0$, we perturb the policy parameters $\theta$ across each of these directions with the scaling $\nu$. The policy is perturbed both along the direction $\theta + \nu\delta_{(i)}$ and away from the direction $\theta - \nu\delta_{(i)}$. Executing this perturbed policy for one episode yields the returns for each direction. Therefore, for $N$ directions, we collect $2N$ returns. We choose the best $N/2$ directions corresponding to the maximum returns. Let $\delta_{(i)}$, $i=1,2,\dots,N/2$ correspond to these best directions in decreasing order of returns. We update $\theta$ as follows:
\begin{align}
\theta = \theta +   \frac{\beta}{\frac{N}{2} * \sigma_R}\sum_{i=1}^{N/2} \left (R(\theta + \nu\delta_{(i)}) - R(\theta - \nu\delta_{(i)})\right  ) \delta_{(i)},
\end{align}
where $\beta>0$ is the step size, and $\sigma_R$ is the standard deviation of the $N/2$ returns obtained. $\theta$ is updated in the matrix $M$, and the gait is tested. This is repeated in every iteration.

\subsection{Training results}
\label{subsec:training}
An online open source physics engine, PyBullet, was used to simulate and train our agent. 
% Our robot model was loaded in the simulator using a Universal Robot Description Format [URDF] file. 
Accurate measurements of link-lengths, moment of inertia's and masses were stored in the URDF file. An asynchronous version of Augmented Random Search with $20$ parallel agents was used to speed up the training on the robot. The reward function ($r$) chosen was
\begin{align}\label{eq:reward}
r = W_{vel} \cdot \Delta x - W_E \cdot \Delta E.
\end{align}
Here $\Delta x$ is the difference between the current and the previous base positions/orientations. This difference changes depending upon the gait. For example, for trotting we seek for positive increments along the $x$-axis, for side-stepping we seek for increments along the $z$-axis, and for turning we seek for increments in yaw. $\Delta E$ is the energy expended to execute the motion in each step.$\Delta E$ is calculated by the formula $E = \tau * v* \Delta t $ where $\tau$ is the joint torque, $v$ is the joint velocity and $\Delta t$ is length of the simulation time step. The energy $E$ is summed over one entire half-step of the robot. $W_{vel}$, $W_E$ are weights corresponding to each of the terms in \eqref{eq:reward}. 

\begin{figure}
\centering
\includegraphics[width = 0.9\linewidth] {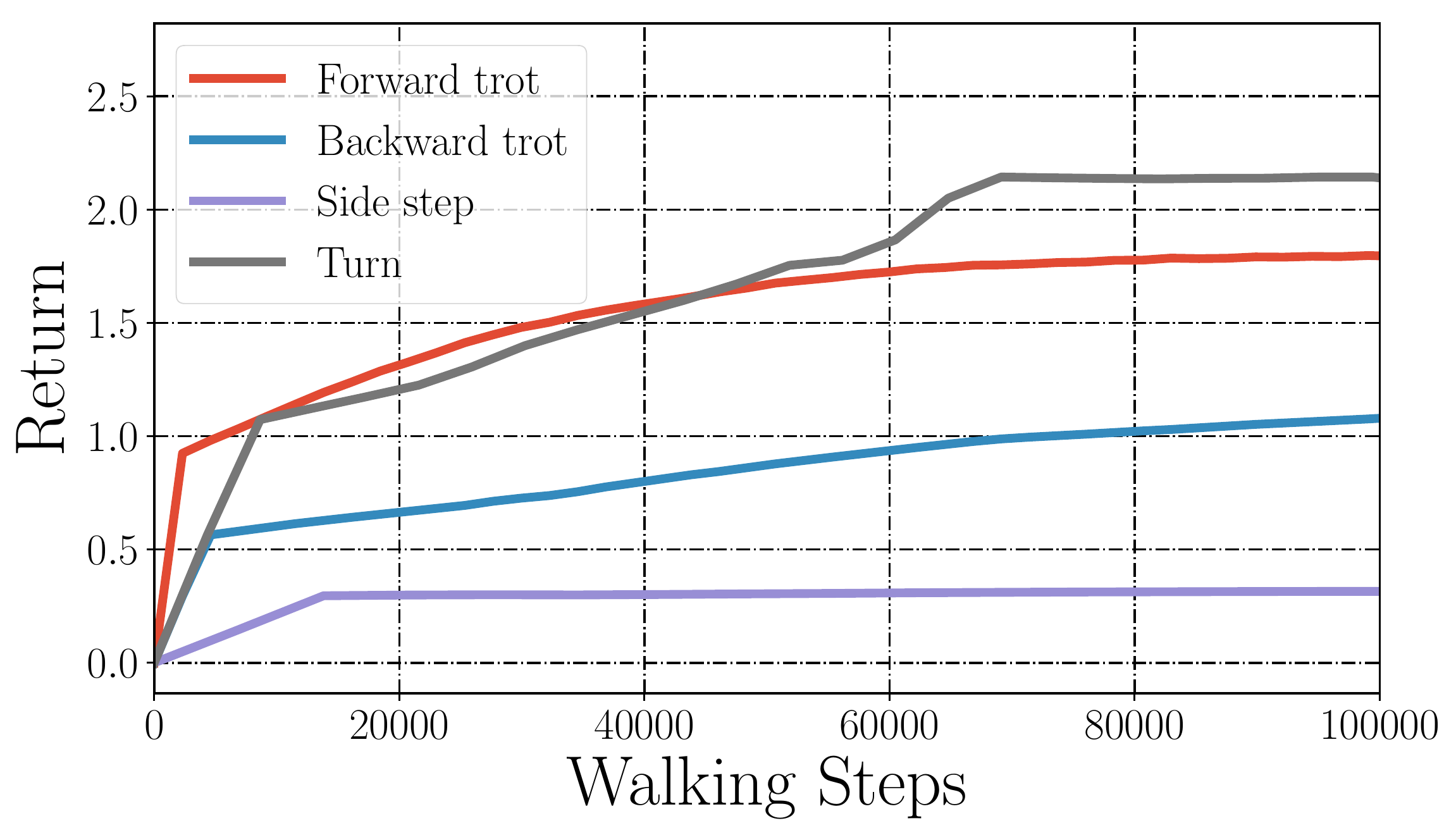}
\caption{Figure showing the training curve for all the four gaits. The returns saturated after about $60000$ gait steps.}
\label{fig:training}
\end{figure}

\begin{table}[]
\begin{center}
\begin{tabular}{|l|l|l|}
\hline
              & Step-Size ($\beta$) & Noise ($\nu$) \\ \hline
Forward trot          & 0.09          & 0.03  \\ \hline
Backward trot & 0.1           & 0.03  \\ \hline
Side step     & 0.1           & 0.03  \\ \hline
Turn     & 0.1           & 0.05  \\ \hline
\end{tabular}
\caption{Table showing the hyperparameters used for the ARS algorithm.}
\label{tab:hyperparameter}
\end{center}
\end{table}

We trained for the following gaits in the pybullet simulator: a) forward trot, b) backward trot, c) side-step and d) turn. These gaits were realized by 1) appropriate orientation of the plane of control points (see Fig. \ref{fig:IKTrajspline}) and 2) by adding a phase offset between the $\alpha$'s of different legs. For forward and backward trotting, the plane was aligned with the $x-y$ axes, while for side-stepping the plane was aligned with the $y-z$ axes. For turning, the plane trajectory was oriented at $+/-45^\circ$ yaw angle w.r.t. the $x$ axis. The planes for the front two legs were facing inwards, and the planes for the back two legs were facing outwards.
The phase offsets for all the four behaviors were $\{ 0, \pi, \pi, 0 \} $, which correspond to front left, front right, back left, and back right respectively. 
Fig. \ref{fig:training} shows the training results for all the four gaits of the robot. The hyper-parameters used for training are shown in Tab. \ref{tab:hyperparameter}. We used $20$ parallel agents for training every gait. The forward and backward trot took $20$ iterations to learn, which corresponds to about $400,000$ simulation time-steps.
Similarly sidestep took $20$ iterations, and turn took $35$ iterations to learn. The training times ranged between $2$-$6$ hours, which are, in fact, on par with the training times obtained with D-RL for Stoch \cite{singla2018realizing}.

\subsection{Training with multiple gaits} We also have preliminary results on training with multiple gaits simultaneously. Multiple agents were spawned in a single environment with different phase differences and were run with the same policy. The new reward was the combined reward of all the agents. To the best of our knowledge ours is the only network capable of exhibiting multiple different quadruped gaits with a single linear policy. The also stands to the testament of the surprising capability of fully linear policies for robotic control tasks first observed in simulation in \cite{mania2018simple} and experimentally in \cite{PMTG}. Our training in all took about 20 iterations to saturate with about 20 parallel agents at a time. So in all it took about 400,000 simulation time-steps. 

\section{Experimental Results}
\label{sec:experiment}
In this section, we will describe the experimental results. We will first describe the methodology used to address the sim-to-real gap.

\subsection{Bridging the sim-to-real gap}
The desired trajectories obtained from the splines are tracked via a joint level PD control law. We observed that, despite having sufficient tracking performances at the joint level, the control points obtained in experiments from the linear policy were not matching with the control points obtained in simulation. This is due to the mismatch between the joint angle trajectories obtained from the simulation and experiment. In simulation, the tracking performance significantly differs from that in the real world. In particular, contact forces, joint and gearbox friction, inaccurate motor models cause the deviation in tracking performance between the simulated and real world data. The policy outputs the control points assuming implicitly a tracking performance observed in simulation and thus the same controllers do not work on the real world robot. 

\begin{figure}[tb]
\centering
\includegraphics[width = 1\linewidth]{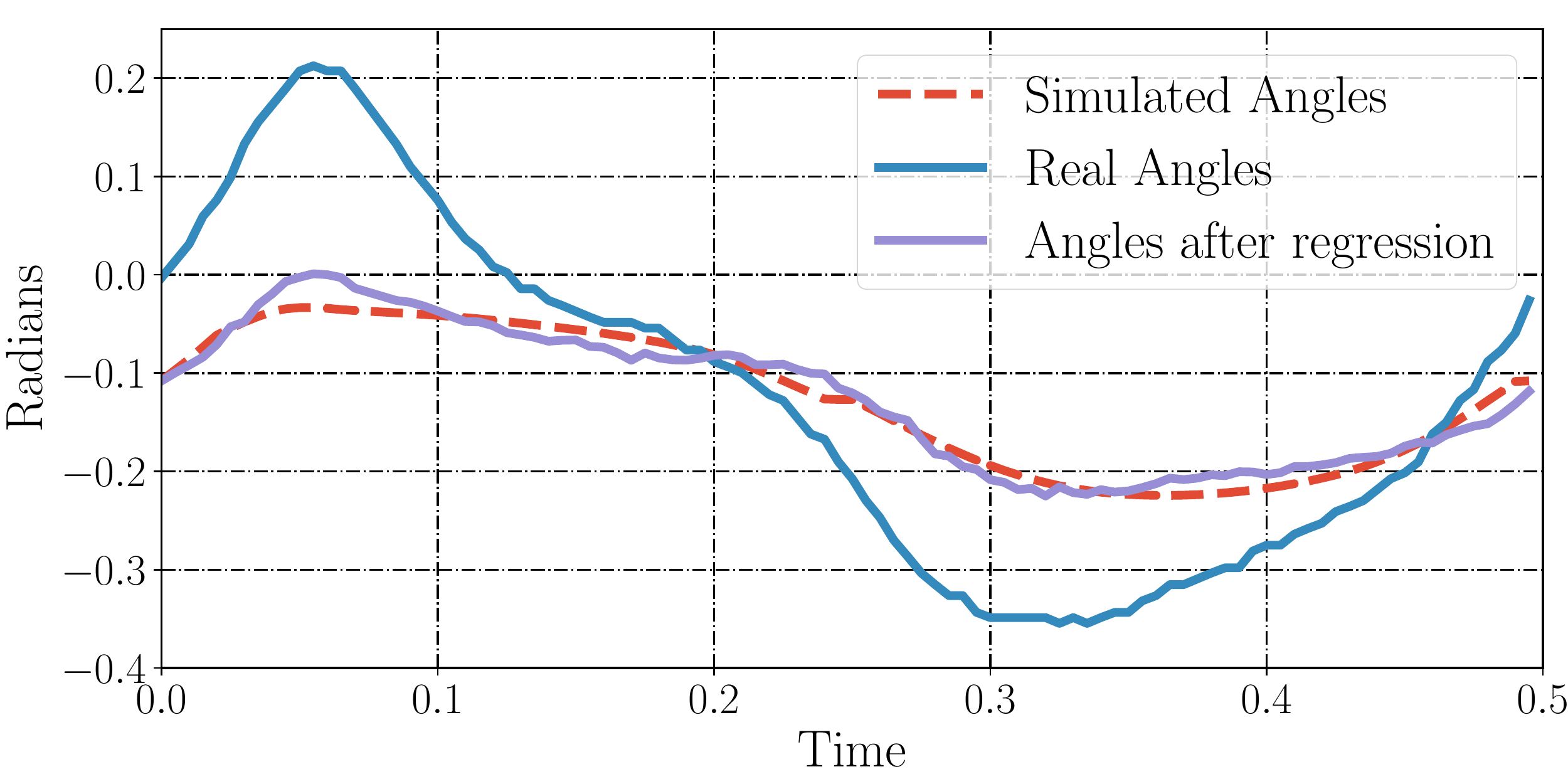}
\includegraphics[width = 1\linewidth]{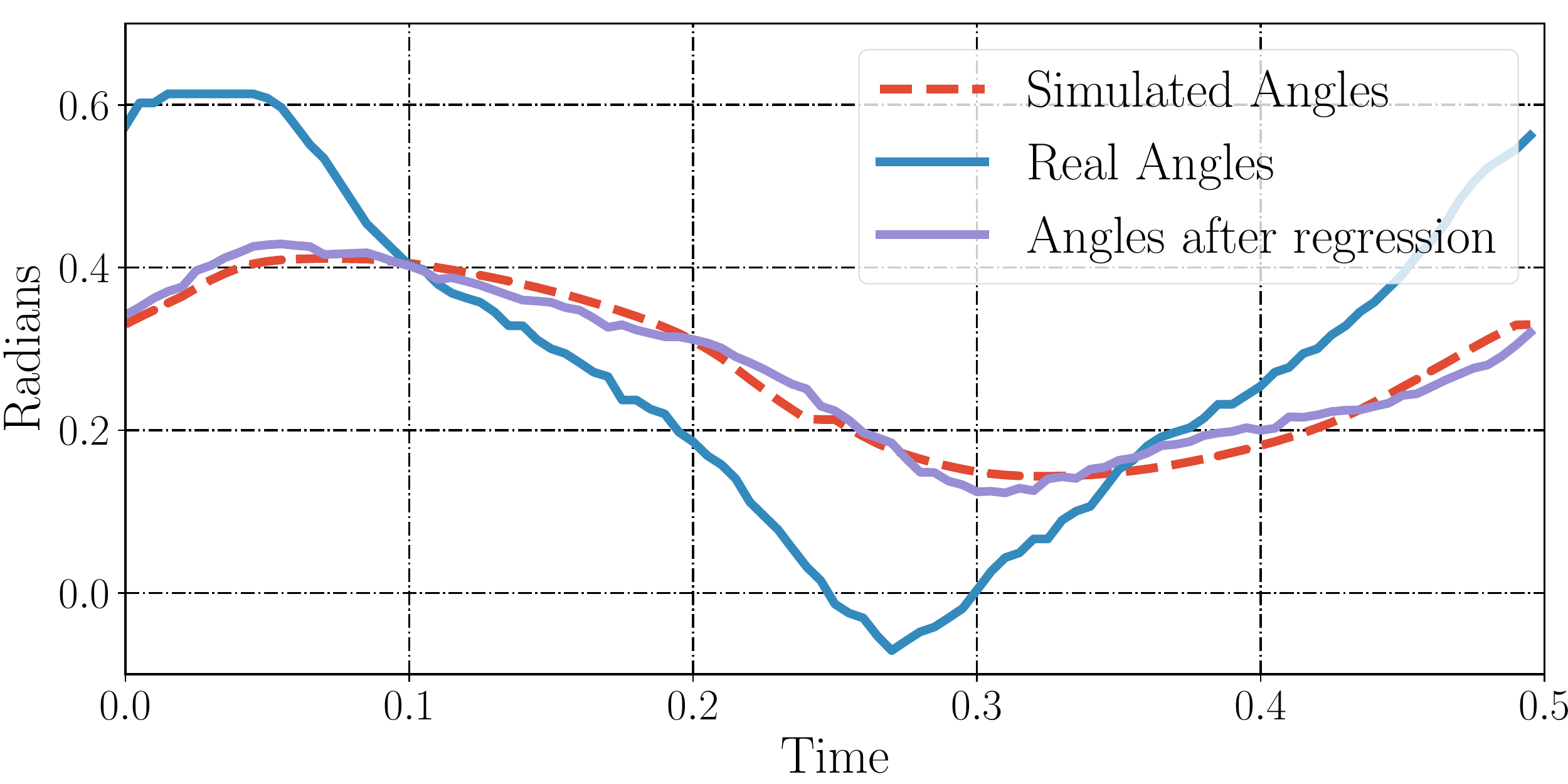}
\caption{Linear regression maps the real world data to the corresponding simulation data. The top figure is for the hip angle, and the bottom is for the knee angle. The dashed red line is the simulated trajectory, and the solid blue line is the experimental trajectory obtained for the same action values (control points) used. The purple line is the experimental trajectory obtained after the transformation.}
\label{fig:LinearRegression}
\end{figure}

To address this sim-to-real problem \cite{google_paper} used domain randomization and a custom motor model. However domain randomization makes the learning unstable and causes early saturation of rewards. Since we have a linear policy, we observed marginal improvements with domain randomization. \cite{Hwangboeaau5872} aimed to bridge this gap by building an accurate motor model via supervised learning. In a similar vein, we addressed the sim-to-real gap by transforming the observed motor angles in the real world in such a way that it matches with the observed motor angles in simulation. 
%This means that since we observed the joint level PD Control law to have satisfactory tracking performance in simulation and the real world, we modified the policy so that it outputs the control points with the implicit assumption of the tracking performance observed in the real world instead of that observed in the simulation. 
We first extracted the control points that our policy outputs when it reaches steady state (~ 5 steps) in simulation. These control points will be used to compare the tracking performance in simulation and experiment. 
%We could have used a circle or any other shape to measure the tracking performance, however we get better results when using control points obtained from our policy as these are the points which will finally be run on the robot. 
In particular, these control points are used to obtain the desired trajectories, which are tracked both in experiment and in simulation. Motor angle data was collected for $5$ steps at a rate of $3$~kHz. Next, a simple weighted linear regression between the experimental and the simulated motor angles collected was used. More weights were given for the stance phase of the legs where the inaccuracies are high due to contact with the ground. Fig. \ref{fig:LinearRegression} shows the comparison between the two data sets obtained. We denote the linear transformation obtained from the regression as $\hat M$. We have the actions obtained from the new policy $\hat \pi$ as
\begin{align}
    \hat \pi(s) : = M \hat M s  + M \bar b,
\end{align}
where $\bar b$ is the offset obtained from the data. The new policy is used to obtain the desirable action values (control points) that match with the control points used in simulation. Thus the new policy outputs the same desired trajectory, for the tracking performance observed in the real world and since the tracking performance of the PD Control Law is good, now the policy works well on the robot, bridging the "sim-to-real" gap. There was a net error of $0.01$~rad between the simulated and experimental state values after performing the weighted linear regression (see Fig. \ref{fig:LinearRegression}). 

\begin{figure}
\centering
\includegraphics[width = \columnwidth] {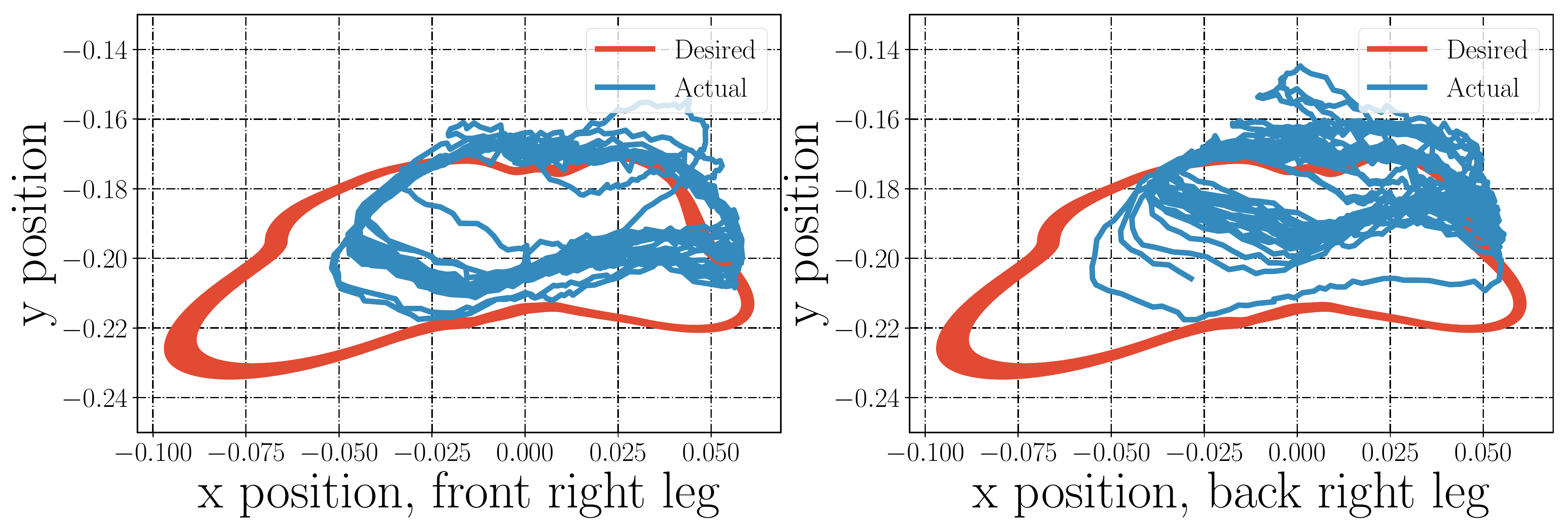} 
\caption{Figure showing the closed operation of the trotting gait on Stoch 2 for $50$ gait steps. It can be observed that the desired trajectories are stable despite the external disturbances.}
\label{fig:limitcycle}
\end{figure}

\begin{figure*}
\centering
\begin{subfigure}[b]{0.48\textwidth}
\includegraphics[width = \columnwidth] {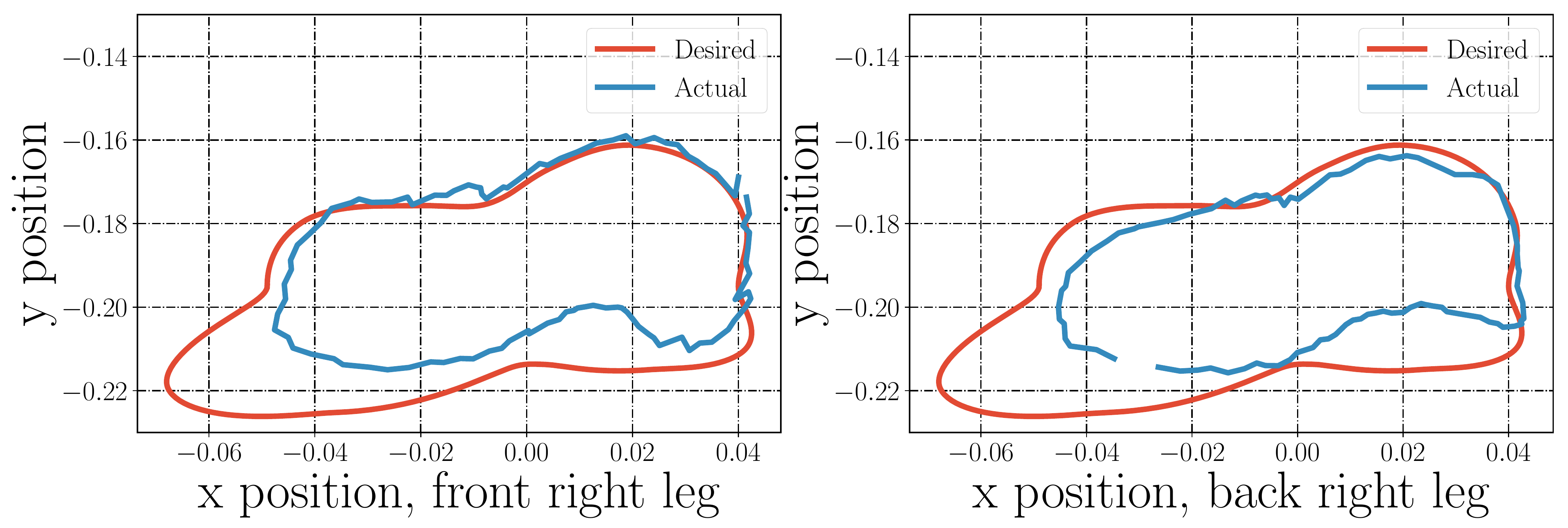}
\caption{Forward trot.}
\vspace{2mm}
\end{subfigure}
\begin{subfigure}[b]{0.48\textwidth}
\includegraphics[width = \columnwidth] {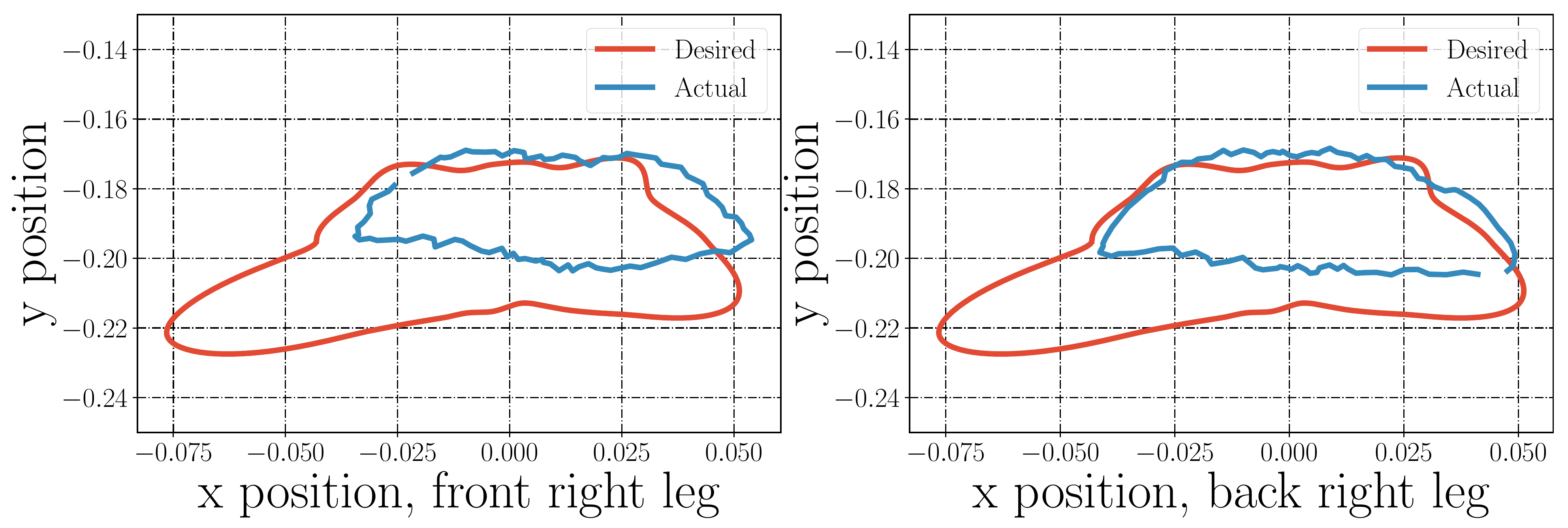}
\caption{Backward trot.}
\vspace{2mm}
\end{subfigure} \\
\begin{subfigure}[b]{0.48\textwidth}
\includegraphics[width = \columnwidth] {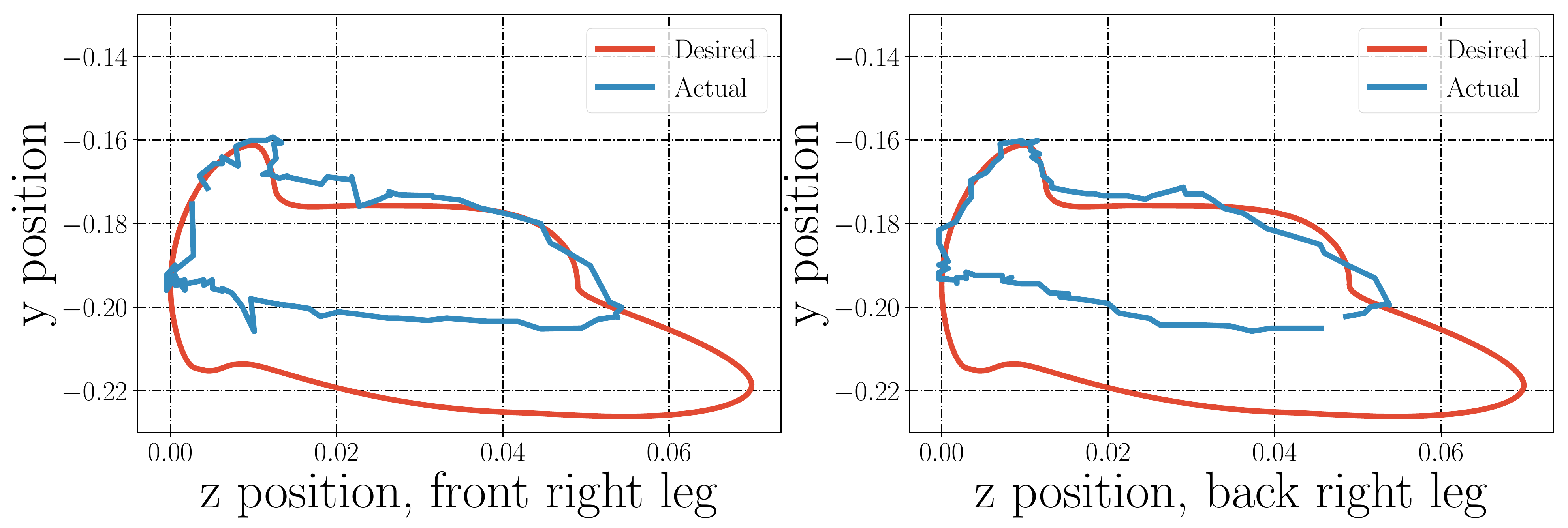} 
\caption{Side-step.}
\end{subfigure} 
\begin{subfigure}[b]{0.48\textwidth}
\includegraphics[width = \columnwidth] {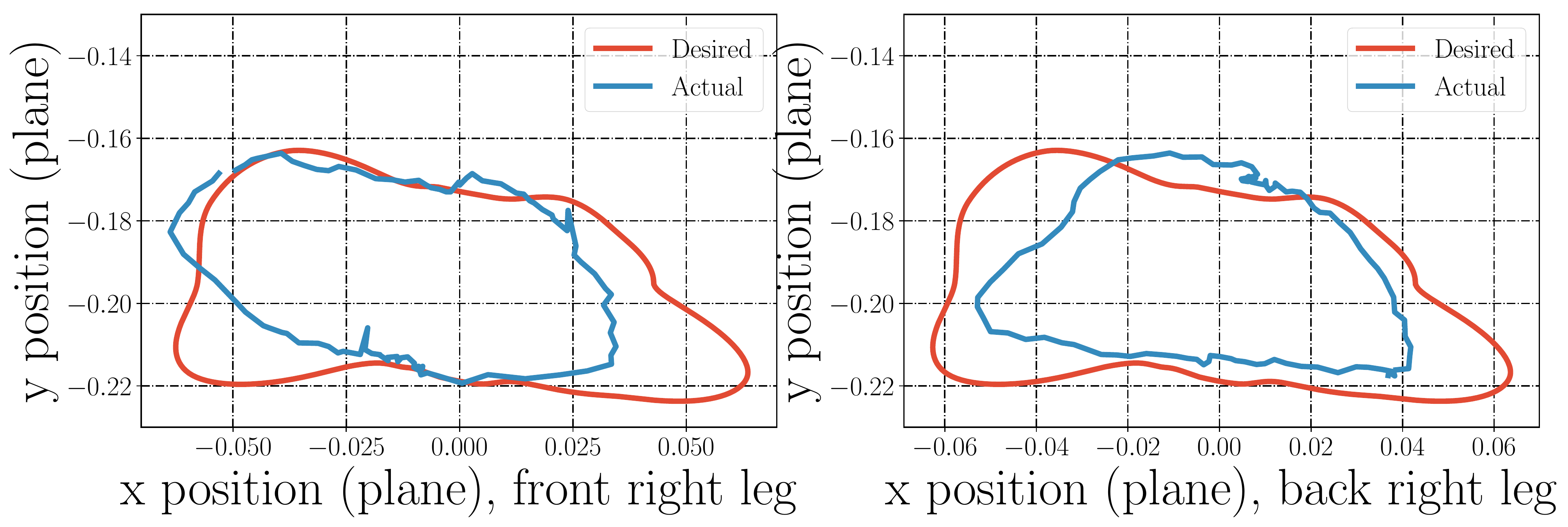} 
\caption{Turn.}
\end{subfigure}
\caption{Figure showing the comparison between the desired (red) and the actual end-foot trajectories for the various gaits tested on the robot. The plots are for one full cycle, i.e., two gait steps. The plots for turning are shown in the plane that is at $45^\circ$ w.r.t. the $x$ axis. Note that all the values are specified in meters.}
\label{fig:end-foottrajectories}
\end{figure*}

\begin{figure*}
\centering
\begin{subfigure}[b]{\textwidth}
\includegraphics[width = \textwidth] {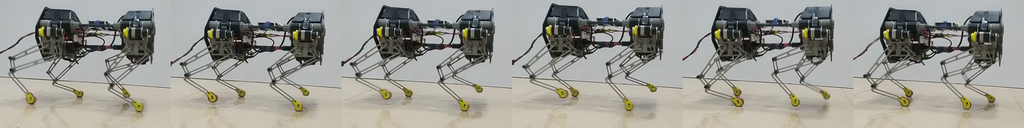}
\caption{Forward trot.}
\vspace{2mm}
\end{subfigure}
\begin{subfigure}[b]{\textwidth}
\includegraphics[width = \textwidth] {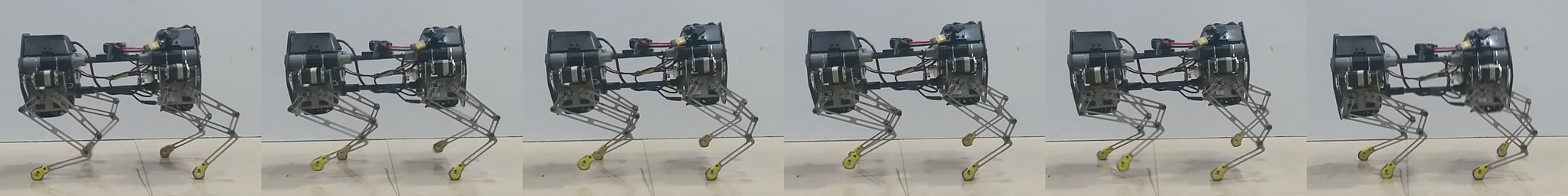} 
\caption{Backward trot.}
\vspace{2mm}
\end{subfigure}
\begin{subfigure}[b]{\textwidth}
\includegraphics[width = \textwidth] {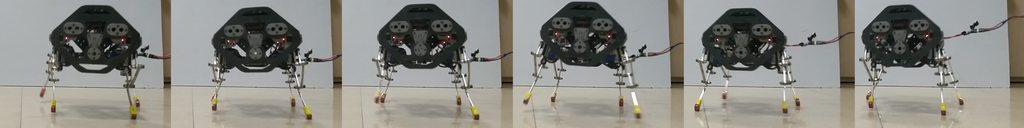} 
\caption{Side-step.}
\vspace{2mm}
\end{subfigure}
\begin{subfigure}[b]{\textwidth}
\includegraphics[width = \textwidth] {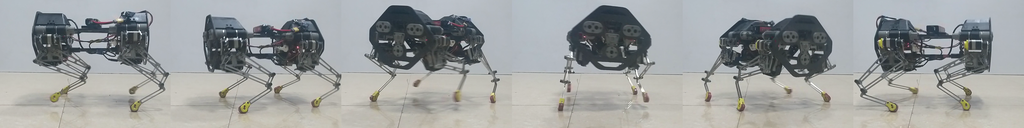} 
\caption{Turn.}
\end{subfigure}
\caption{Figure showing the sequence of tiles for all the four gaits tested on Stoch 2. The tiles for forward-backward trot, and side-step are for one gait step, while the tiles for turn are for approximately eight gait steps.}
\label{fig:walkingtiles}
\end{figure*}

\subsection{Results}

We tested all the four gaits on Stoch 2 experimentally. As mentioned previously, the states observed at the end of every gait step were transformed to yield more desirable state values, which were then used to obtain the desired trajectories for the next step. 
Fig. \ref{fig:end-foottrajectories} shows the comparison between the desired and actual end-foot trajectories of the robot, and Fig. \ref{fig:walkingtiles} shows the corresponding gait tiles for all the four gaits.
While each of these gaits were obtained from training separately, the transition from one type of gait to another was achieved by using a low pass filter (see \cite{dhaivatdesigndevelopment} for more details).
Video results showing all the four gaits are provided in the submission.

\subsection{Empirical analysis of limit cycle behaviour}
Fig. \ref{fig:limitcycle} shows the robustness of the trotting gait to external disturbances. The data recorded was for fifty gait steps. The video submission shows the disturbances applied on the robot, and the desired trajectories converging to a limit cycle.
Experiments show that joint angles achieve self-correcting behaviors, which implies that a linear policy is able to calculate the control points that reject the disturbances.

\section{Conclusion}

We successfully demonstrated multiple types of locomotion gaits learned via Augmented Random Search (ARS) in the custom built quadruped robot Stoch $2$. Gaits were first tested in a simulation environment, and then deployed in Stoch $2$ experimentally. Linear policy was used to determine the trajectory for each gait step. These types of policies are easy to deploy on hardware, and, at the same time, do not require large compute capability, unlike the Deep Neural Network (DNN) based control algorithms existing in literature. It is worth noting that training times with ARS are comparable with D-RL with the same compute platform. Future work will involve learning and testing more gaits on a diverse set of terrains like slopes and stairs.

%%%%%%%%%%%%%%%%%%%%%%%%%%%%%%%%%%%%%%%%%%%%%%%%%%%%%%%%%%%%%%%%%%%%%%%%%%%%%%%%

%%%%%%%%%%%%%%%%%%%%%%%%%%%%%%%%%%%%%%%%%%%%%%%%%%%%%%%%%%%%%%%%%%%%%%%%%%%%%%%%



\section*{ACKNOWLEDGMENT}
The authors would like to thank Alok Rawat for assisting with the design and manufacturing of Stoch 2.

%%%%%%%%%%%%%%%%%%%%%%%%%%%%%%%%%%%%%%%%%%%%%%%%%%%%%%%%%%%%%%%%%%%%%%%%%%%%%%%%

% \bibliographystyle{unsrtnat}
\clearpage
\bibliographystyle{IEEEtran}
\bibliography{references}

\begin{thebibliography}{10}
\providecommand{\url}[1]{#1}
\csname url@samestyle\endcsname
\providecommand{\newblock}{\relax}
\providecommand{\bibinfo}[2]{#2}
\providecommand{\BIBentrySTDinterwordspacing}{\spaceskip=0pt\relax}
\providecommand{\BIBentryALTinterwordstretchfactor}{4}
\providecommand{\BIBentryALTinterwordspacing}{\spaceskip=\fontdimen2\font plus
\BIBentryALTinterwordstretchfactor\fontdimen3\font minus
  \fontdimen4\font\relax}
\providecommand{\BIBforeignlanguage}[2]{{%
\expandafter\ifx\csname l@#1\endcsname\relax
\typeout{** WARNING: IEEEtran.bst: No hyphenation pattern has been}%
\typeout{** loaded for the language `#1'. Using the pattern for}%
\typeout{** the default language instead.}%
\else
\language=\csname l@#1\endcsname
\fi
#2}}
\providecommand{\BIBdecl}{\relax}
\BIBdecl

\bibitem{raibert1986legged}
M.~H. Raibert, ``Legged robots,'' \emph{Communications of the ACM}, vol.~29,
  no.~6, pp. 499--514, 1986.

\bibitem{vukobratovic2004zero}
M.~Vukobratovi{\'c} and B.~Borovac, ``Zero-moment point—thirty five years of
  its life,'' \emph{International journal of humanoid robotics}, vol.~1,
  no.~01, pp. 157--173, 2004.

\bibitem{hzd_grizzle}
E.~R. Westervelt, J.~W. Grizzle, and D.~E. Koditschek, ``Hybrid zero dynamics
  of planar biped walkers,'' \emph{IEEE Transactions on Automatic Control},
  vol.~48, no.~1, pp. 42--56, Jan 2003.

\bibitem{ijspeert2008central}
A.~J. Ijspeert, ``Central pattern generators for locomotion control in animals
  and robots: a review,'' \emph{Neural networks}, vol.~21, no.~4, pp. 642--653,
  2008.

\bibitem{kober2013reinforcement}
J.~Kober, J.~A. Bagnell, and J.~Peters, ``Reinforcement learning in robotics: A
  survey,'' \emph{The International Journal of Robotics Research}, vol.~32,
  no.~11, pp. 1238--1274, 2013.

\bibitem{google_paper}
\BIBentryALTinterwordspacing
J.~Tan, T.~Zhang, E.~Coumans, A.~Iscen, Y.~Bai, D.~Hafner, S.~Bohez, and
  V.~Vanhoucke, ``Sim-to-real: Learning agile locomotion for quadruped
  robots,'' \emph{CoRR}, vol. abs/1804.10332, 2018. [Online]. Available:
  \url{http://arxiv.org/abs/1804.10332}
\BIBentrySTDinterwordspacing

\bibitem{Hwangboeaau5872}
\BIBentryALTinterwordspacing
J.~Hwangbo, J.~Lee, A.~Dosovitskiy, D.~Bellicoso, V.~Tsounis, V.~Koltun, and
  M.~Hutter, ``Learning agile and dynamic motor skills for legged robots,''
  \emph{Science Robotics}, vol.~4, no.~26, 2019. [Online]. Available:
  \url{https://robotics.sciencemag.org/content/4/26/eaau5872}
\BIBentrySTDinterwordspacing

\bibitem{xie2019iterative}
Z.~Xie, P.~Clary, J.~Dao, P.~Morais, J.~Hurst, and M.~van~de Panne, ``Iterative
  reinforcement learning based design of dynamic locomotion skills for
  cassie,'' \emph{arXiv preprint arXiv:1903.09537}, 2019.

\bibitem{Hornby1999Autonomous}
G.~Hornby, M.~Fujita, S.~Takamura, T.~Yamamoto, and O.~Hanagata, ``Autonomous
  evolution of gaits with the sony quadruped robot,'' 1999.

\bibitem{quinlan2003techniques}
M.~J. Quinlan, S.~K. Chalup, R.~H. Middleton \emph{et~al.}, ``Techniques for
  improving vision and locomotion on the sony aibo robot.''

\bibitem{kohl2004policy}
N.~Kohl and P.~Stone, ``Policy gradient reinforcement learning for fast
  quadrupedal locomotion,'' in \emph{Robotics and Automation, 2004.
  Proceedings. ICRA'04. 2004 IEEE International Conference on}, vol.~3.\hskip
  1em plus 0.5em minus 0.4em\relax IEEE, pp. 2619--2624.

\bibitem{tedrake2004stochastic}
R.~Tedrake, T.~W. Zhang, and H.~S. Seung, ``Stochastic policy gradient
  reinforcement learning on a simple 3d biped,'' in \emph{Intelligent Robots
  and Systems, 2004.(IROS 2004). Proceedings. 2004 IEEE/RSJ International
  Conference on}, vol.~3.\hskip 1em plus 0.5em minus 0.4em\relax IEEE, pp.
  2849--2854.

\bibitem{Chalup2007Machine}
S.~K. Chalup, C.~L. Murch, and M.~J. Quinlan, ``Machine learning with aibo
  robots in the four-legged league of robocup,'' \emph{IEEE Transactions on
  Systems, Man, and Cybernetics, Part C (Applications and Reviews)}, vol.~37,
  no.~3, pp. 297--310, 2007.

\bibitem{mania2018simple}
H.~Mania, A.~Guy, and B.~Recht, ``Simple random search of static linear
  policies is competitive for reinforcement learning,'' in \emph{Advances in
  Neural Information Processing Systems}, 2018, pp. 1800--1809.

\bibitem{PMTG}
A.~Iscen, K.~Caluwaerts, J.~Tan, T.~Zhang, E.~Coumans, V.~Sindhwani, and
  V.~Vanhoucke, ``Policies modulating trajectory generators,'' in
  \emph{Conference on Robot Learning}, 2018, pp. 916--926.

\bibitem{jain2019hierarchical}
D.~Jain, A.~Iscen, and K.~Caluwaerts, ``Hierarchical reinforcement learning for
  quadruped locomotion,'' \emph{arXiv preprint arXiv:1905.08926}, 2019.

\bibitem{singla2018realizing}
A.~Singla, S.~Bhattacharya, D.~Dholakiya, S.~Bhatnagar, A.~Ghosal, B.~Amrutur,
  and S.~Kolathaya, ``Realizing learned quadruped locomotion behaviors through
  kinematic motion primitives,'' \emph{arXiv preprint arXiv:1810.03842}, 2018.

\bibitem{dhaivatdesigndevelopment}
\BIBentryALTinterwordspacing
D.~Dholakiya, S.~Bhattacharya, A.~Gunalan, A.~Singla, S.~Bhatnagar, B.~Amrutur,
  A.~Ghosal, and S.~Kolathaya, ``Design, development and experimental
  realization of a quadrupedal research platform: Stoch,'' \emph{CoRR}, vol.
  abs/1901.00697, 2019. [Online]. Available:
  \url{http://arxiv.org/abs/1901.00697}
\BIBentrySTDinterwordspacing

\bibitem{action_space}
\BIBentryALTinterwordspacing
X.~B. Peng and M.~van~de Panne, ``Learning locomotion skills using deeprl: Does
  the choice of action space matter?'' \emph{CoRR}, vol. abs/1611.01055, 2016.
  [Online]. Available: \url{http://arxiv.org/abs/1611.01055}
\BIBentrySTDinterwordspacing

\bibitem{PPO}
\BIBentryALTinterwordspacing
J.~Schulman, F.~Wolski, P.~Dhariwal, A.~Radford, and O.~Klimov, ``Proximal
  policy optimization algorithms,'' \emph{CoRR}, vol. abs/1707.06347, 2017.
  [Online]. Available: \url{http://arxiv.org/abs/1707.06347}
\BIBentrySTDinterwordspacing

\bibitem{TRPO}
\BIBentryALTinterwordspacing
J.~Schulman, S.~Levine, P.~Moritz, M.~I. Jordan, and P.~Abbeel, ``Trust region
  policy optimization,'' \emph{CoRR}, vol. abs/1502.05477, 2015. [Online].
  Available: \url{http://arxiv.org/abs/1502.05477}
\BIBentrySTDinterwordspacing

\end{thebibliography}

\end{document}